\def\year{2021}\relax
\documentclass[letterpaper]{article} % DO NOT CHANGE THIS

\usepackage{times}
\usepackage{graphicx} % more modern
\usepackage{epsfig} % less modern
\usepackage{subfigure} 

\usepackage{natbib}

\usepackage{algorithm}
\usepackage{algorithmic}

\usepackage{helvet}
\usepackage{courier}

\usepackage{makecell}
\usepackage{graphicx}
\usepackage{subfigure}
\usepackage{color}
\usepackage{amsfonts}
\usepackage{amsmath}
\usepackage{amssymb}

\def\ie{\textit{i.e.}}
\def\eg{\textit{e.g.}}

\DeclareMathAlphabet\mathbfcal{OMS}{cmsy}{b}{n}
\def\0{{\bf 0}}
\def\1{{\bf 1}}

\usepackage{booktabs}

\usepackage{aaai21}  % DO NOT CHANGE THIS
\usepackage{times}  % DO NOT CHANGE THIS
\usepackage{helvet} % DO NOT CHANGE THIS
\usepackage{courier}  % DO NOT CHANGE THIS
\usepackage[hyphens]{url}  % DO NOT CHANGE THIS
\usepackage{graphicx} % DO NOT CHANGE THIS
\usepackage{natbib}  % DO NOT CHANGE THIS AND DO NOT ADD ANY OPTIONS TO IT
\usepackage{caption} % DO NOT CHANGE THIS AND DO NOT ADD ANY OPTIONS TO IT
\usepackage{multirow}
\usepackage{footnote}
\usepackage{xcolor}
 
\title{Conditional Automated Channel Pruning for Deep Neural Networks}
\author{
    Yixin Liu\textsuperscript{$\dag$}, Yong Guo\textsuperscript{$\dag$}, Zichang Liu\textsuperscript{$\dag$}, Haohua Liu\textsuperscript{$\dag$}, Jingjie Zhang\textsuperscript{$\dag$}, \\Zejun Chen\textsuperscript{$\dag$}, Jing Liu\textsuperscript{$\ddag$}, Jian Chen\textsuperscript{$\dag$}\thanks{Corresponding author.}
    \\
}
\affiliations{
    \textsuperscript{$\dag$}South China University of Technology,
    \textsuperscript{$\ddag$}Monash University
    \{seyixinliu, guo.yong, se\_liuzc, seforlawa\_mail, sezjj, seczj722\}@mail.scut.edu.cn,\\ liujing\_95@outlook.com, ellachen@scut.edu.cn
}
\begin{document}

\maketitle

\begin{abstract}
{
Model compression aims to reduce the redundancy of deep networks to obtain compact models. Recently, channel pruning has become one of the predominant compression methods to deploy deep models on resource-constrained devices. Most channel pruning methods often use a fixed compression rate for all the layers of the model, which, however, may not be optimal. To address this issue, given a target compression rate for the whole model, one can search for the optimal compression rate for each layer. Nevertheless, these methods perform channel pruning for a specific target compression rate. When we consider multiple compression rates, they have to repeat the channel pruning process multiple times, which is very inefficient yet unnecessary. To address this issue, we propose a Conditional Automated Channel Pruning (CACP) method to obtain the compressed models with different compression rates through single channel pruning process. To this end, we develop a conditional model that takes an arbitrary compression rate as input and outputs the corresponding compressed model. In the experiments, the resultant models with different compression rates consistently outperform the models compressed by existing methods with a channel pruning process for each target compression rate.
}
\end{abstract}

\section{Introduction}
Deep Neural Networks (DNNs) has achieved great success in many tasks, e.g., image classification, face recognition, and video analysis. 
However, deep models often contain a large number of parameters and require high computational resources.As a result, it is hard to apply deep learning methods to resource-constrained devices. To address this, model compression has been an effective way to reduce redundancy of deep networks~\cite{zhuang2018discrimination,guo2019nat}. 

\begin{figure}[t]
    \centering
    \includegraphics[width=1\columnwidth]{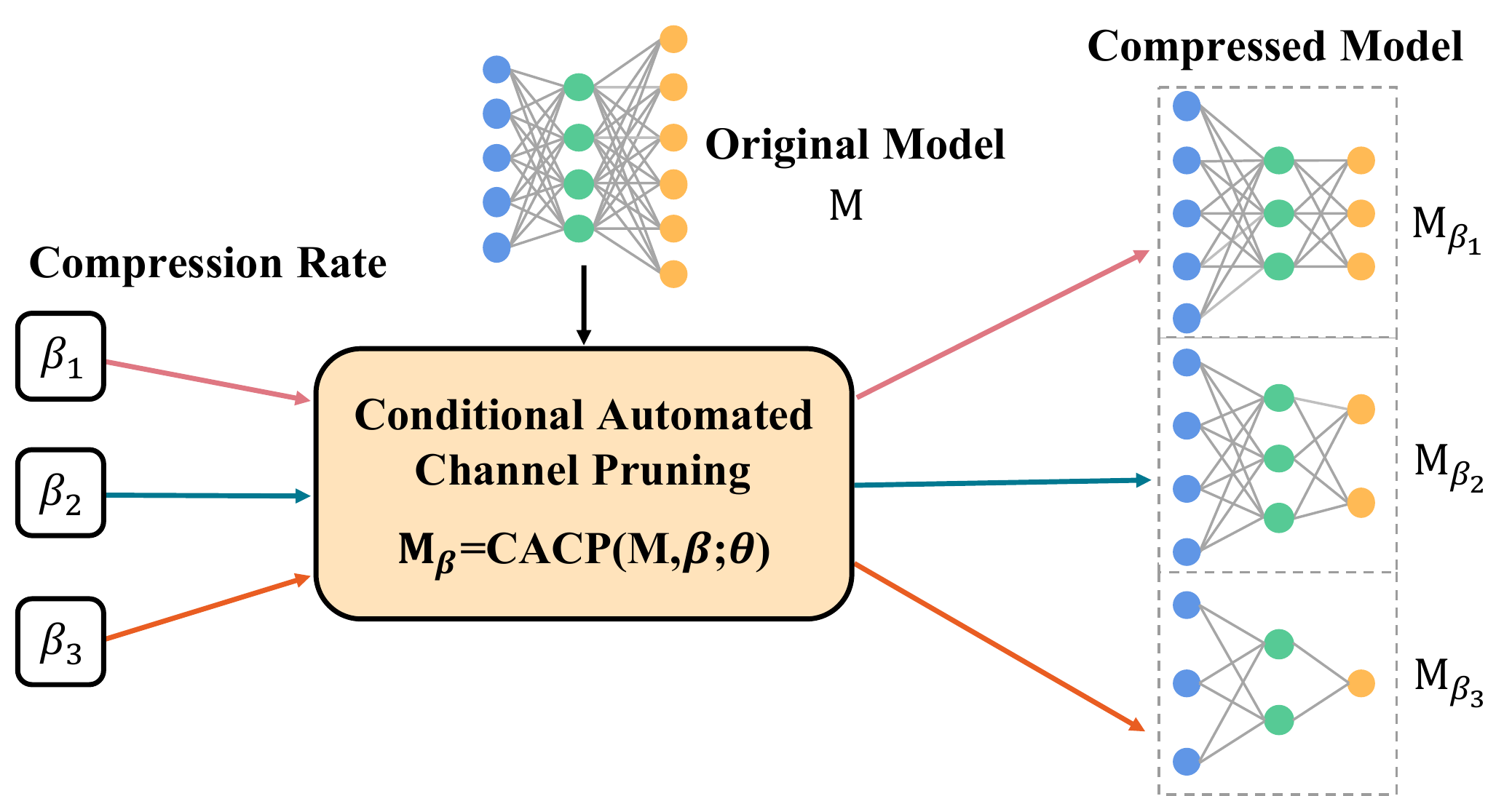}
    \caption{The overview of the proposed CACP. Our CACP takes a pretrained model $M$ and a target compression rate as inputs and outputs the compressed model that satisfies the considered target compression rate.}
    \label{fig:overview}
\end{figure}

Recently, channel pruning has been one of the predominant approaches for deep model compression. Specifically, channel pruning aims to remove the redundant channels of the layers in a deep network.
To obtain the models with the desired compactness, most methods apply a fixed compression rate to all the layers of the deep model~\cite{zhuang2018discrimination,he2019filter}. However, not all layers have the same amount of redundancy. Thus, pruning the same proportion of channels for all the layer may be suboptimal. To address this issue, \citet{he2018amc} propose an automatic pruning method AMC that searches for the optimal compression rate for each layer. However, these methods only perform channel pruning for a specific target compression rate. When we consider different compression rates, they have to repeat the channel pruning for each compression rate, which is very time-consuming and labor-intensive.

To address the above issue, we seek to train a single model to obtain the compressed models with different target compression rates simultaneously (See Figure~\ref{fig:overview}).
To this end, we propose a Conditional Automated Channel Pruning (CACP) method that takes the target compression rate as the condition to obtain the compressed model satisfying this condition. Extensive experiments show that the resultant models obtained by our CACP significantly outperform the compressed models by the considered baseline methods.

\section{Conditional Automated Channel Pruning}

Existing channel pruning methods may either apply a fixed compression rate to all the layers or search for the optimal compression rate for each layer to satisfy the overall target rate. However, when we consider different target compression rates, we have to repeat the channel pruning process to obtain the models satisfying these target rates, which is very inefficient yet unnecessary.

To address this issue, we propose a Conditional Automated Channel Pruning (CACP) method that automatically compresses models with different target compression rates. In this sense, we only train the CACP model once to obtain the models with different computational cost (\eg, FLOPs) simultaneously. To achieve this, we treat the target compression rate as a condition and train a conditional model to obtain the compressed models satisfying the desired conditions.
Given a pretrained model $M$ and any arbitrary target compression rate $\beta$, we seek to obtain the compressed model by 
$M_\beta = {\rm CACP}(M,\beta;\theta)$, where $\theta$ denotes the learnable parameters of the CACP model. 
To enable CACP to compress models under different target compression rates, we train the model by maximizing the expected reward over a distribution of compression rate, \ie, $\beta {\sim} p(\cdot)$. In this paper, we assume $p(\cdot)$ to be a uniformly discrete distribution. During training, we use the validation accuracy as the reward $R(\cdot)$. Thus, the objective can be formulated as

\begin{equation}
    \max_{\theta} ~\mathbb{E}_{\beta \sim p(\cdot)} \left[  R \left(M_\beta \right) \right].
\end{equation}

However, directly obtaining the compressed models is non-trivial. Instead, we seek to determine the compression rate for each layer and perform channel selection to obtain the compressed models.
Following~\cite{he2018amc}, we use reinforcement learning to search for the optimal compression rate in a layer-wise manner. As for channel selection, we use L1 norm to measure the importance of channels~\cite{han2015learning} and prune the unimportant channels based on the compression rates. 
Note that we have to limit the compression rate when we find that the resultant model cannot reach the target compression rate even though we remove all the channels of the following layers.
Let $C_l$ be the computational cost of this layer, and $D_{l}^{(\beta)}$ be the lower bound of the computational cost that should be reduced for the $l$-th layer to achieve the compression rate $\beta$.
To ensure that we can obtain the model satisfying the overall compression rate, the compression rate for the $l$-th layer becomes
\begin{equation}\label{eq:compression_rate}
    \alpha_l^{(\beta)} = \max \Big( f(S_l, \beta; \theta), ~{D_{l}^{(\beta)}}{\big/}{C_l} \Big),
\end{equation}

where $S_l$ denotes the state/features of this layer, $f(\cdot)$ denotes the function that determines the optimal compression rate for layer $l$ under the target compression rate $\beta$. Let $\alpha_{\rm max}$ be the maximum possible compression rate for the following layers, $C_{\rm all}$ be the overall cost of the whole model, $C_{\rm reduced}$ be the total amount of reduced cost in the previous layers, and $C_{\rm rest}$ be the amount of the remaining cost in the following layers. Thus, the minimum computation cost that should be reduced for layer $l$ becomes

\begin{equation}\label{eq:minimum_cost}
    D_{l}^{(\beta)} = \beta \cdot C_{\rm all} - \alpha_{\rm max} \cdot C_{\rm rest} - C_{\rm reduced}.
\end{equation}
In this way, based on Eqns.~(\ref{eq:compression_rate}) and~(\ref{eq:minimum_cost}), we can guarantee that the compressed model would satisfy the target compression rate. 
During inference, based on a well-learned CACP model, we only need to feed in a specific target compression rate as a condition to obtain the optimal compression rates for all the layers under this condition. Then, we perform channel selection to obtain the desired compressed model with the desired target compression rate (See Figure~\ref{fig:overview}).

\section{Experiments}
In this section, we empirically evaluate the proposed CACP method on CIFAR-10. Several state-of-the-art methods
are adopted as the baselines, including
SFP~\cite{he2018soft}, DCP~\cite{zhuang2018discrimination}, and AMC~\cite{he2018amc}. 
From Table~\ref{cmac_acc}, the models obtained by CACP significantly outperform the considered baseline methods with all compression rates. It is worth noting that all the resultant models are obtained through a single channel pruning process, which is essentially different from existing methods. 
\begin{table}[h]
\centering
\caption{Comparisons of the compressed ResNet-56 models on CIFAR-10. ``-'' denotes the results that are not reported.}
\label{cmac_acc}
\resizebox{1\linewidth}{!}
{
\begin{tabular}{c|c|c|c|c}
\toprule
Compression Rate      & Method              & Acc. (\%) & \#FLOPs $\downarrow$ (\%)  & \#Params. $\downarrow$ (\%)   \\ \hline
0 & Baseline & 93.80 & 0 & 0 \\
\hline
\multirow{3}{*}{0.3} 
& SFP  &         93.59  &  28.4  &  - \\ 
& AMC                 &     93.75          &  31.1 & 19.5 \\ 
& {CACP (Ours)} &      \textbf{93.98}      & 30.2 & 28.2\\ 
\hline
\multirow{4}{*}{0.5}  
& SFP                 &         92.57     &  52.6 & -  \\ 
& DCP &     93.77       &  50.6  & 49.7   \\ 
& AMC                 &     93.57        &  49.9 & 44.6  \\ 
& {CACP (Ours)} &    \textbf{93.84}    & 50.3 &  45.9 \\ 
\hline
\multirow{3}{*}{0.7} & DCP &     92.98       &  68.4  & 68.3    \\ 
& AMC                 &      92.61      &   69.9 &   69.8\\ 
& {CACP (Ours)} &   \textbf{93.13}    &  69.9  &   71.2 \\ 
\bottomrule
\end{tabular}
}
\end{table}
\section{Conclusion}
In this paper, we have proposed a Conditional Automated Channel Pruning method (CACP) that obtains the compressed models with different target compression rates through a single channel pruning process. Specifically, we treat the target compression rate as a condition and train a conditional pruning model to compress deep networks. Extensive experiments show that our compressed models with different compression rates consistently outperform the considered baseline methods. 
\bibliography{aaai21}

\begin{thebibliography}{6}
\providecommand{\natexlab}[1]{#1}
\providecommand{\url}[1]{\texttt{#1}}
\providecommand{\urlprefix}{URL }
\expandafter\ifx\csname urlstyle\endcsname\relax
  \providecommand{\doi}[1]{doi:\discretionary{}{}{}#1}\else
  \providecommand{\doi}{doi:\discretionary{}{}{}\begingroup
  \urlstyle{rm}\Url}\fi

\bibitem[{Guo et~al.(2019)Guo, Zheng, Tan, Chen, Chen, Zhao, and
  Huang}]{guo2019nat}
Guo, Y.; Zheng, Y.; Tan, M.; Chen, Q.; Chen, J.; Zhao, P.; and Huang, J. 2019.
\newblock Nat: Neural architecture transformer for accurate and compact
  architectures.
\newblock In \emph{NeurIPS}.

\bibitem[{Han et~al.(2015)Han, Pool, Tran, and Dally}]{han2015learning}
Han, S.; Pool, J.; Tran, J.; and Dally, W. 2015.
\newblock Learning both weights and connections for efficient neural network.
\newblock In \emph{NeurIPS}.

\bibitem[{He et~al.(2018{\natexlab{a}})He, Kang, Dong, Fu, and
  Yang}]{he2018soft}
He, Y.; Kang, G.; Dong, X.; Fu, Y.; and Yang, Y. 2018{\natexlab{a}}.
\newblock Soft Filter Pruning for Accelerating Deep Convolutional Neural
  Networks.
\newblock In \emph{IJCAI}.

\bibitem[{He et~al.(2018{\natexlab{b}})He, Lin, Liu, Wang, Li, and
  Han}]{he2018amc}
He, Y.; Lin, J.; Liu, Z.; Wang, H.; Li, L.-J.; and Han, S. 2018{\natexlab{b}}.
\newblock Amc: Automl for model compression and acceleration on mobile devices.
\newblock In \emph{ECCV}.

\bibitem[{He et~al.(2019)He, Liu, Wang, Hu, and Yang}]{he2019filter}
He, Y.; Liu, P.; Wang, Z.; Hu, Z.; and Yang, Y. 2019.
\newblock Filter pruning via geometric median for deep convolutional neural
  networks acceleration.
\newblock In \emph{CVPR}, 4340--4349.

\bibitem[{Zhuang et~al.(2018)Zhuang, Tan, Zhuang, Liu, Guo, Wu, Huang, and
  Zhu}]{zhuang2018discrimination}
Zhuang, Z.; Tan, M.; Zhuang, B.; Liu, J.; Guo, Y.; Wu, Q.; Huang, J.; and Zhu,
  J. 2018.
\newblock Discrimination-aware channel pruning for deep neural networks.
\newblock In \emph{NeurIPS}.

\end{thebibliography}
\bigskip
\end{document}